\documentclass[lettersize,journal]{IEEEtran}
\usepackage{amsmath,amsfonts}
\usepackage{algorithmic}
\usepackage{algorithm}
\usepackage{array}
\usepackage[caption=false,font=normalsize,labelfont=sf,textfont=sf]{subfig}
\usepackage{textcomp}
\usepackage{stfloats}
\usepackage{url}
\usepackage{verbatim}
\usepackage{graphicx}
\usepackage{cite}
\usepackage{multirow}
\usepackage{booktabs}
\usepackage{color,xcolor}

\newcommand{\figref}[1]{Fig. \ref{#1}}
\newcommand{\tabref}[1]{Tab. \ref{#1}}

\begin{document}

\title{Rethinking Lightweight Salient Object Detection via Network Depth-Width Tradeoff}

\author{
    Jia Li, \emph{Senior Member, IEEE}\textsuperscript{},
    Shengye Qiao\textsuperscript{},
    Zhirui Zhao\textsuperscript{},
    Chenxi Xie\textsuperscript{} and
    Xiaowu Chen, \emph{Senior Member, IEEE}\textsuperscript{},
    Changqun Xia\textsuperscript{}
    \footnote{Correspondence should be addressed to Changqun Xia and Jia Li. URL: http://cvteam.net.}
    \\
        
\thanks{
Jia Li is with the State Key Laboratory of Virtual Reality Technology and Systems, School of Computer Science and Engineering, Beihang University, Beijing, 100191, China, and also with Peng Cheng Laboratory, Shenzhen, 518000, China.

Shengye Qiao, Zhirui Zhao and Chenxi Xie are with the State Key Laboratory of Virtual Reality Technology and Systems, School of Computer Science and Engineering, Beihang University, Beijing, 100191, China.

Xiaowu Chen is with the State Key Laboratory of Virtual Reality Technology and Systems, School of Computer Science and Engineering, Beihang University, Beijing, 100191, China, and also with Peng Cheng Laboratory, Shenzhen, 518000, China.

Changqun Xia is with Peng Cheng Laboratory, Shenzhen, 518000, China.

Corresponding author: Changqun Xia(E-mail: xiachq@pcl.ac.cn)

A preliminary version of this work has appeared in ACM MM 2021\cite{zhao2021complementary}

}
}
\markboth{SUBMISSION TO IEEE TRANSACTIONS ON IMAGE PROCESSING}%
{Shell \MakeLowercase{\textit{et al.}}: A Sample Article Using IEEEtran.cls for IEEE Journals}

\IEEEpubid{}

\maketitle

\begin{abstract}
  Existing salient object detection methods often adopt deeper and wider networks for better performance, resulting in heavy computational burden and slow inference speed. This inspires us to rethink saliency detection to achieve a favorable balance between efficiency and accuracy. To this end, we design a lightweight framework while maintaining satisfying competitive accuracy. Specifically, we propose a novel trilateral decoder framework by decoupling the U-shape structure into three complementary branches, which are devised to confront the dilution of semantic context, loss of spatial structure and absence of boundary detail, respectively. Along with the fusion of three branches, the coarse segmentation results are gradually refined in structure details and boundary quality. Without adding additional learnable parameters, we further propose Scale-Adaptive Pooling Module to obtain multi-scale receptive filed. In particular, on the premise of inheriting this framework, we rethink the relationship among accuracy, parameters and speed via network depth-width tradeoff. With these insightful considerations, we comprehensively design shallower and narrower models to explore the maximum potential of lightweight SOD. Our models are purposed for different application environments: 1) a tiny version CTD-S (1.7M, 125FPS) for resource constrained devices, 2) a fast version CTD-M (12.6M, 158FPS) for speed-demanding scenarios, 3) a standard version CTD-L (26.5M, 84FPS) for high-performance platforms. Extensive experiments validate the superiority of our method, which achieves better efficiency-accuracy balance across five benchmarks.
\end{abstract}


\section{Introduction}
\IEEEPARstart{I}{n} recent years, salient object detection (SOD) \cite{borji2019salient, wang2021salient} has made great progress with the development of Deep Neural Networks (DNNs). As an efficient preprocessing technique, SOD plays an important role in many downstream computer vision tasks, such as image retrieval\cite{gao2015database}, visual tracking\cite{hong2015online} and semantic segmentation \cite{haralick1985image}. 

\begin{figure}[t]
  \centering
  \includegraphics[width=1.0\columnwidth]{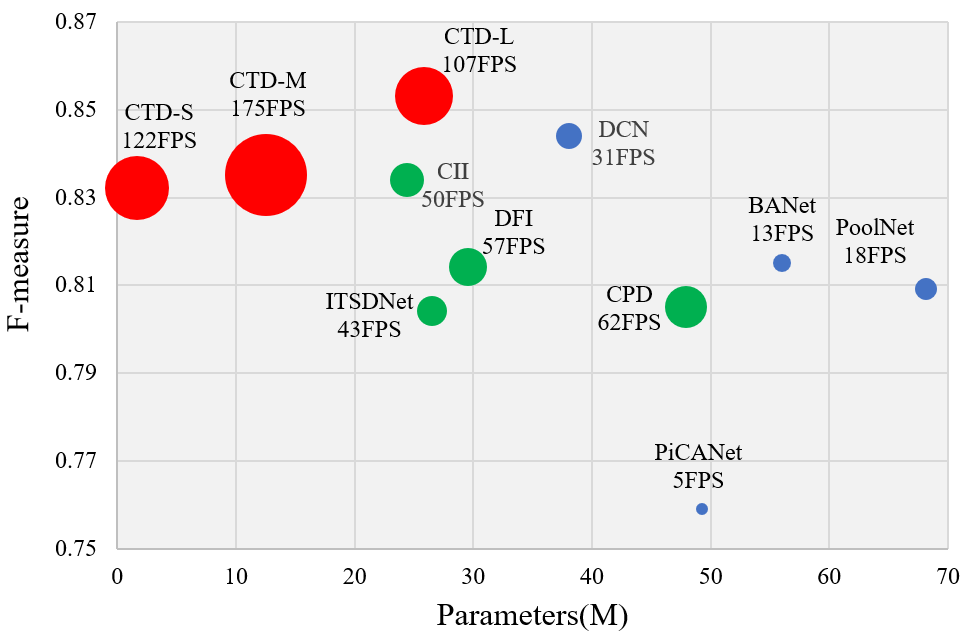} 
  \caption{Comparisons of our proposed CTDNet with other ResNet-based SOD models in accuracy, parameters and speed. We calculate $mF_\beta$ on DUTS-TE dataset as an example. The green circles represent real-time SOD methods, blue circles represent non-real-time SOD methods, and red circles represent our method. The size of circle indicates speed and larger circle indicates faster speed.}
  \label{fig1}
\end{figure}

Many SOD methods have greatly benefited from very deep and wide models and achieves remarkable results. For example, BANet\cite{su2019banet} and DCN\cite{wu2021decomposition} reach high accuracy compared to other models as shown in \figref{fig1}. However, the success comes at a price of heavy computation burden and slow running speed. These models increase the network depth and width by adjusting the number of layers and channels, which brings tremendous parameters and calculations. Taking these factors into consideration, a question arises: \emph{Is shallow and narrow network able to achieve comparable performance of large counterparts?}

Some researches start to answer this question and try to compromise between efficiency and accuracy, such as GCPANet\cite{chen2020global} and ITSDNet\cite{zhou2020interactive}. These methods can lead to improvements in efficiency but still adopt strong classification models pretrained on ImageNet(e.g., ResNet-50 and ResNet-101) as backbone, which tend to bring about a large amount of parameters. For example, ITSDNet contains about 27M parameters and the ones of GCPANet are approaching 67M. Besides, since SOD does not care about the category of objects, the features extracted by overly deep networks may be overqualified. To this end, some methods have adopted efficient backbones when designing networks and shown great potential for establishing highly competitive models with fewer parameters and faster speed. For example, EDN\cite{wu2022edn} using MobileNet-V2\cite{sandler2018mobilenetv2} as the backbone only contains 1.8M parameters with fewer channels, and it has comparable performance due to inverted block and depth-wise separable convolutions. CII\cite{liu2021rethinking} using ResNet-18\cite{he2016deep} as the backbone is a representative shallow network with a depth of just 17 layers, which significantly improves the processing speed. Therefore, making full use of advantages of these efficient backbones to solve SOD problem is a way worth exploring and trying.

Besides backbone, the complexity of the framework itself is also an important factor affecting the efficiency of the model. It is worth noting that most state-of-the-art SOD methods belong to the U-shape structure\cite{ronneberger2015u}, where the input image is gradually down-sampled to extract multi-scale features and then up-sampled to recover the resolution with skip connection, as shown in \figref{fig3}(a). Except for strong backbone models in the encoder, these methods design complex structures in the decoder to refine saliency maps, but inevitably introduce additional parameters. In addition, there are several problems for the U-shaped structure. First, spatial information is seriously lost during the process of frequent down-sampling and cannot be perfectly recovered by integrating the hierarchical features\cite{yu2018bisenet}, which leads to incomplete structure details. Second, semantic and global context information may be gradually diluted in the top-down path, which leads to inaccurate object location. Third, boundary information is also ignored, which leads to poor boundary quality. To address these challenges, it is necessary to design a new and effective method to deal with the dilution of semantic context, loss of spatial structure and absence of boundary detail.

\begin{figure}[t]
  \centering
  \includegraphics[width=1.0\columnwidth]{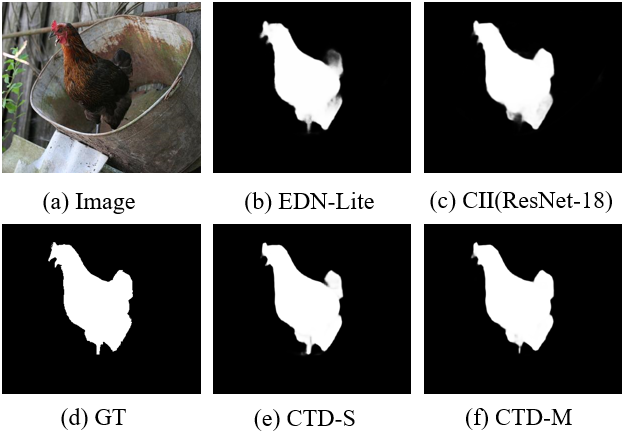} 
  \caption{The saliency maps of the same image from EDN-Lite, CII(ResNet-18), CTD-S and CTD-M}
  \label{fig2}
\end{figure}

With the above considerations in mind, we propose a lightweight framework that performs competitively against other large counterparts. Instead of conventional backbone models, we rely on efficient backbones for feature extraction. We take advantage of these backbones more effectively to encode hierarchical features from low-level, mid-level to high-level, which can fully meet the requirements of SOD task at minimal computation cost. In addition to the encoder, we propose a novel Complementary Trilateral Decoder (CTD) that decouples the U-shape structure into three branches: Semantic Path, Spatial Path and Boundary Path. As illustrated in \figref{fig3}(b), these three branches are derived from different levels of the shared encoder and complementary to each other. Along with the fusion of three branches, the coarse segmentation results are refined in structure details and boundary quality. Specifically, Semantic Path is designed to capture rich semantic context and global context of high-level features, which can form initial coarse saliency maps with accurate locations of salient objects. Spatial Path is introduced to preserve more spatial details of mid-level features and then is combined with Semantic Path by the proposed Cross Aggregation Module (CAM), which can produce relative fine saliency maps with precise structures of salient objects. Boundary Path is utilized to extract salient boundary contour by fusing low-level local features and high-level location features with an explicit edge supervision. Finally, the output of CAM and Boundary Path are merged by the proposed Boundary Refinement Module (BRM) to further refine boundary, which can generate final finer saliency maps with clear boundaries of salient objects. Besides, although semantic category features are unnecessary for SOD task, the multi-scale representation ability of high-level features is indispensable. Therefore, we take the advantages of pooling operation and propose a novel structure named Scale-Adaptive Pooling (SAP) to capture multi-scale context information from multiple receptive fields without additional learnable parameters. 

More importantly, on the basis of inheriting this framework, we rethink the relationship among accuracy, parameters and speed according to the different depth and width characteristics of efficient backbones, thus discover more potential of lightweight SOD. To facilitate the practical application in different environments, we provide three versions: 1) a tiny version CTD-S (1.7M, 105FPS) for resource-constrained devices, 2) a fast version CTD-M (12.6M, 160FPS) for speed-demanding scenarios, 3) a standard version CTD-L (25.9M, 100FPS) for high-performance platforms. Extensive experiments on five popular SOD datasets demonstrate the generality and superiority of our method, which achieves better balance between efficiency and accuracy.

In general, the main contributions of our work are summarized as follows:

1) We rethink SOD task from the perspective of network depth and width, and then design lightweight and efficient saliency detection models with shallow and narrow networks, which can perform surprisingly well against other large counterparts with less parameters and higher efficiency.

2) We propose a novel Complementary Trilateral Decoder (CTD) framework that decouples the U-shape structure into three branches: Semantic Path, Spatial Path and Boundary Path. The Cross Aggregation Module (CAM) and Boundary Refinement Module (BRM) are constructed to gradually merge these three complementary branches according to “coarse-fine-finer” strategy, which significantly improves the region accuracy and boundary quality.

3) We take the advantages of pooling operation to enhance the multi-scale representation ability of high-level features and propose a novel Scale-Adaptive Pooling (SAP) structure to obtain multi-scale receptive filed without increasing parameters. 

4) We provide three versions for different application environments and conduct extensive experiments on 5 benchmarks to demonstrate the generality and superiority of our method, which achieves a favorable trade-off between efficiency and accuracy.

Compared with the conference version of CTDNet, we optimize the parameters and accuracy from the perspective of width and depth under the premise of being lightweight, and propose three versions of the model based on complementary trilateral decoder framework: CTD-S, CTD-M and CTD-L. Parameters and speeds of different magnitudes in these three models can be more comprehensively adapted to various devices and scenarios. In addition to the original three branches and the corresponding fusion module, we introduce the SAP structure in the decoder. Meanwhile, the decoding process of the three models is also significantly different from the conference version, which will be explained in detail in the methodology section.

\begin{figure}[t]
  \centering
  \includegraphics[width=1.0\columnwidth]{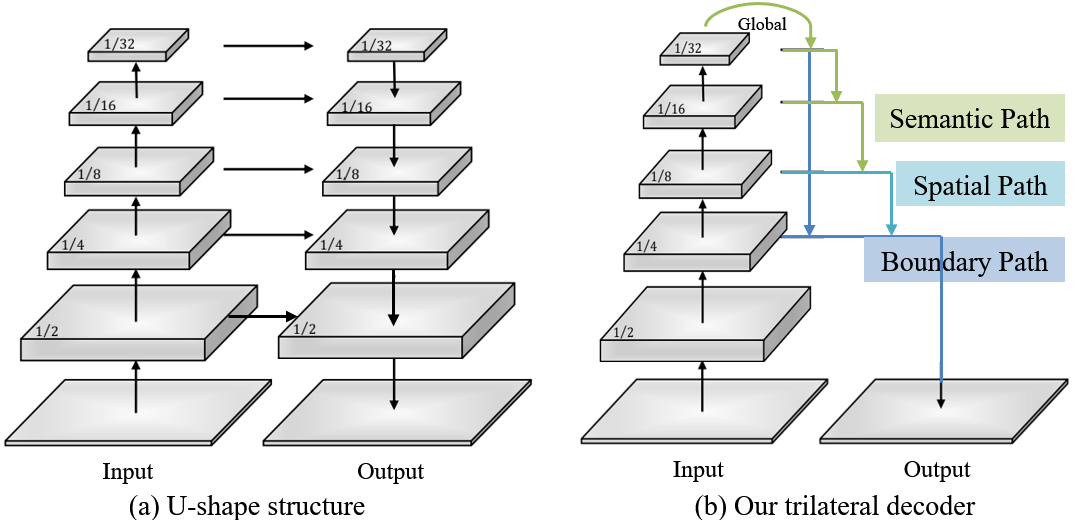} 
  \caption{The U-shape structure gradually recovers the spatial information by leveraging lateral connections and topdown path in the decoder part, while our trilateral decoder separately treats semantic context, spatial detail and boundary information with three branches.}
  \label{fig3}
\end{figure}

\section{Related work}
\subsection{Accurate SOD}
Traditional SOD methods mostly rely on heuristic priors (e.g., color, texture and contrast) to generate saliency maps. However, these hand-crafted features can hardly capture high-level information, which are not robust enough for complex scenarios\cite{xia2017and}. In recent years, many FCN-based methods have achieved remarkable progress thanks to its powerful representation capability. As one of the most representative networks, the U-shape structure has been widely followed for accurate saliency detection. PiCANet \cite{liu2018picanet} proposed a pixel-wise contextual attention network to learn informative context locations for each pixel. BMPM \cite{zhang2018bi} designed a bi-directional message passing model for better feature selection and integration. MINet \cite{pang2020multi} focused on scale variation and class imbalance challenges by utilizing multi-level and multi-scale features. PSGLoss \cite{yang2021progressive} introduced an additional progressive self-guided loss and multi-scale feature aggregation module with branch-wise attention to detect salient objects completely and effectively. Some methods introduce an additional boundary-aware branch or a boundary-aware loss function for fine object boundaries. C2SNet \cite{li2018contour} presented a contour-to-saliency transferring model that predicts contours and saliency maps simultaneously. BASNet \cite{qin2019basnet} proposed a boundary-aware model and designed hybrid loss to make full use of boundary information. EGNet \cite{zhao2019egnet} focused on the complementary information modeling between salient edge and salient object to improve boundaries and localization. BANet \cite{su2019banet} designed a boundary-aware model with successive dilation from the perspective of selectivity and invariance. AFNet \cite{feng2019attentive} proposed a multi-scale attentive feedback model and Boundary-Enhanced Loss to predict salient objects with entire structure and exquisite boundaries. PurNet \cite{li2021salient} utilized promotion and rectification attention to purify salient objects and introduced a structural similarity loss to restore the complex or fine structures of salient objects. In addition, some models are used for accurate segmentation of different scenes, such as automatic driving\cite{su2021exploring} and 360 panoramic scenes\cite{li2019distortion}. However, these methods have brought huge parameters and model complexity for better performance, resulting in slow inference speed.

\begin{figure*}[t]
\centering
\includegraphics[width=1.0\textwidth]{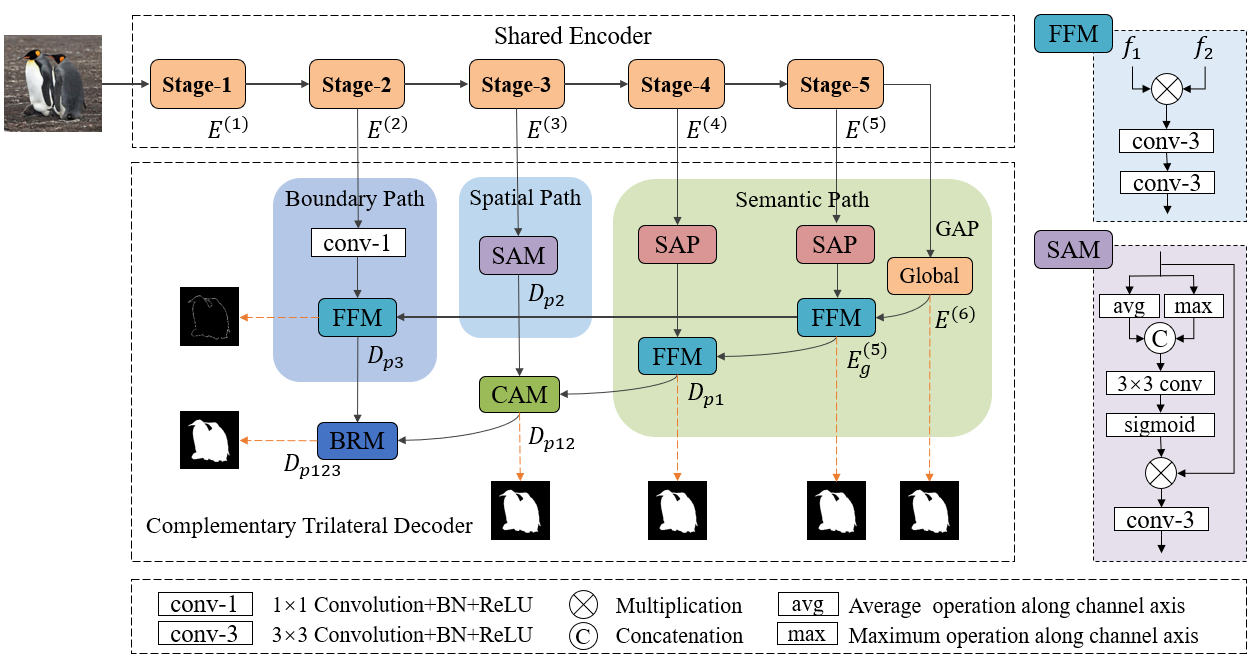} 
\caption{The framework of our proposed Complementary Trilateral Decoder (CTD) Network taking CTD-M as an example with three branches: Semantic Path, Spatial Path and Boundary Path, which treats semantic context, spatial detail and boundary information separately in the decoder part. The three parts share the same encoder and are derived from different stages of the encoder. The three branches are complementary to each other and we design three specific fusion modules to gradually merge them according to “coarse-fine-fine” strategy}
\label{fig4}
\end{figure*}

\subsection{Efficient SOD}
SOD serves as a preprocessing technique for many downstream vision tasks, so accuracy and efficiency are both important factors when building a SOD network. Recently, some methods have been proposed for accurate and fast saliency detection. For example, RAS \cite{chen2018reverse} predicted a global saliency map and proposed reverse attention to guide side-output residual features recursively. PoolNet \cite{liu2019simple} fully exploited the pooling operations based on the FPN structure for real-time salient object detection. CPD \cite{wu2019cascaded} discarded features of shallow layers for acceleration and proposed a cascade partial decoder that utilizes attention mechanism to refine high-level features. ITSDNet \cite{zhou2020interactive} proposed an interactive two-stream decoder to explore multiple cues, including saliency, contour and their correlation. DCN \cite{wu2021decomposition} leveraged skeleton and edge information to model interiors and boundaries of salient objects together in decomposition and completion framework. CII \cite{liu2021rethinking} designed a centralized information interaction strategy to encode the cross-scale information into the learnable filters and modeled information with relative global calibration module. In addition to these resnet-based methods, there are methods using lightweight backbone to improve efficiency. For example, EDNLite \cite{wu2022edn}, adopting MobileNet V2 as backbone model, designed an extremely-downsampled block to learn a global view of the whole image and constructed a scale-correlated pyramid convolution for effective feature fusion in the decoder. In order to design and achieve the ultimate lightweight model, SAMNet \cite{liu2021samnet} abandoned the wildly used pre-trained Convolutional Neural Network (CNN)\cite{shelhamer2017fully} backbones thoroughly and rebuilt a lightweight encoder-decoder architecture with stereoscopically attentive multi-scale module, which adopted stereoscopic attention mechanism for effective and efficient multi-scale learning. Although smaller and faster than the previous large models, these methods cannot achieve comparable performance.

\subsection{Encoder-Decoder Structure}
Recently, many researches follow the U-shape structure to effectively combine low-level and high-level features for saliency detection.TDBU \cite{wang2019iterative} learnt top-down and bottom-up saliency inference in a cooperative and iterative manner. MINet \cite{pang2020multi} focused on scale variation and class imbalance challenges by utilizing multi-level and multi-scale feature information. DASNet \cite{zhao2020depth} proposed a depth-aware framework to improve the segmentation performance with depth constraints. PFSNet \cite{ma2021pyramidal} proposed to aggregate adjacent feature nodes in pairs through layer by layer shrinkage, which can fuse details and semantics effectively, and discard interference information. However, these methods based on U-shape structure have brought  high model complexity to achieve better performance, resulting in slow inference speed.

Different from these methods, we design a less complex structure via network depth-width tradeoff. Apart from adopting efficient backbones as encoder,  we introduce complementary trilateral decoder with cross aggregation module, boundary refinement module and scale-adaptive pooling structure to maintain satisfying competitive accuracy while ensuring the lightness of the model.

\section{METHODOLOGY}
\subsection{Motivation and Framework}
As mentioned above, conventional SOD methods adopt strong backbone models to encode deep semantic information (i.e., category), which inherits a large number of parameters and calculations. However, SOD is a category-insensitive task that focuses on segmenting the salient objects or regions in an image, so it may be unnecessary to extract rich category features from very deep networks. In addition, SOD serves as a preprocessing technique for many downstream vision tasks, so accuracy and efficiency are both important factors when building a SOD network. To this end, we explore the potential of shallow and narrow models, and then specially design a lightweight framework while maintaining satisfying accuracy. The key to the design concept lies in the following aspects: 1) We rely on shallow and narrow backbone models to encode hierarchical features fast while reducing computational burden; 2) We prune the number of feature channels for less redundant information and faster inference speed; 3) We take the advantages of pooling operation to enhance multi-scale representation ability of model; 4) We propose a novel framework for accurate and efficient saliency detection.

Taking into account the drawbacks of the U-shape structure mentioned above, we propose a novel framework that treats semantic context, spatial detail and boundary contour separately in the decoder. The whole architecture of our proposed Complementary Trilateral Decoder (CTD) is shown in \figref{fig4}.

Existing backbones, like ResNet and MobileNet, can encode images in multiple stages and output corresponding feature maps, which can be expressed as  $\{E^1,E^2,E^3,E^4,E^5\}$ for convenience. In order to make full use of the feature maps of different stages in the decoding process, we decouple the decoder of the U-shape structure into three branches: Semantic Path, Spatial Path and Boundary Path, which are devised to confront the dilution of semantic context, loss of spatial structure and absence of boundary detail, respectively. These three branches share the same encoder and are derived from different levels of the encoder. In order to make full use of the characteristics and complementarity of these three branches, we propose the Cross Aggregation Module (CAM) and Boundary Refinement Module (BRM) to gradually merge them according to “coarse-fine-finer” strategy, which significantly improves the region accuracy and boundary quality. Considering the multi-scale representation of high-level features and efficient computation simultaneously, we take the advantages of pooling operation and propose a novel Scale-Adaptive Pooling (SAP) structure. Benefiting from the unique framework and these lightweight designs, our method can generate more accurate segmentation results with fewer parameters and faster speed.

\begin{figure*}[t]
  \centering
  \includegraphics[width=1.0\textwidth]{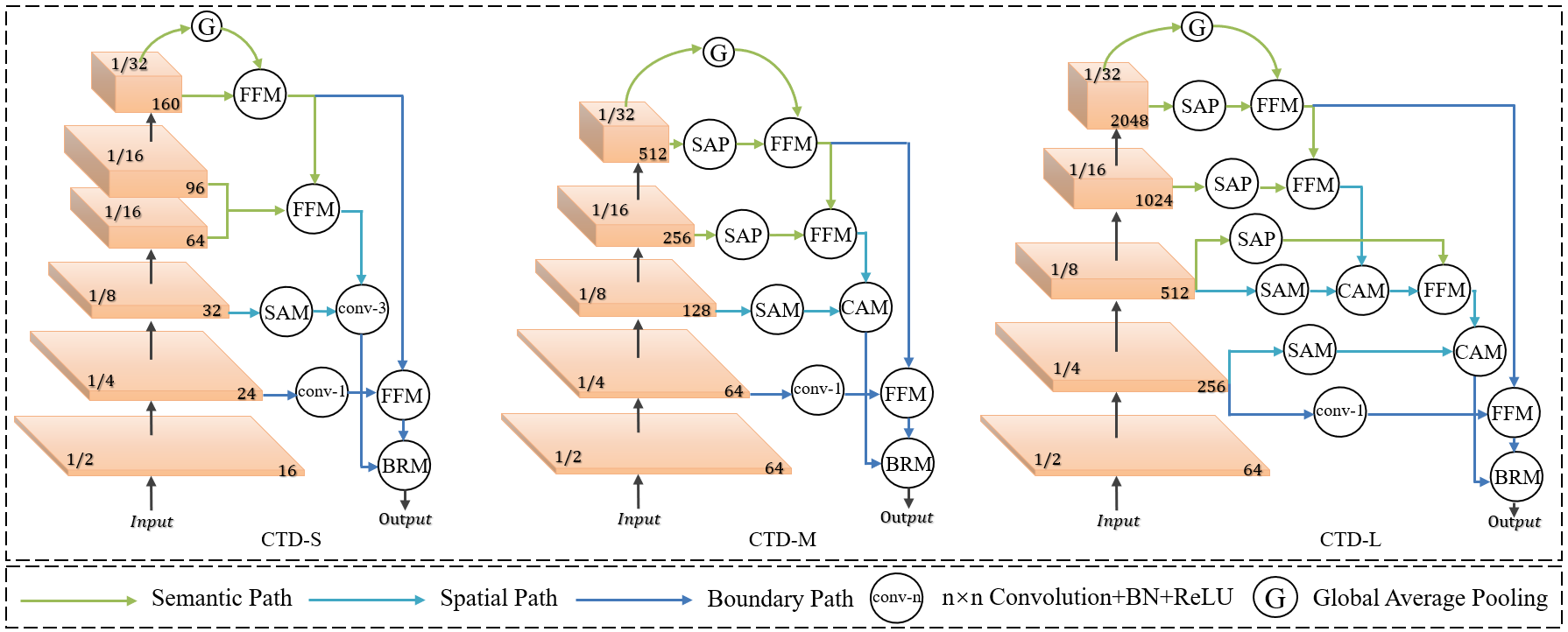} 
  \caption{Comparison of framework details on CTD-S, CTD-M and CTD-L, which adopt MobileNet-V2, ResNet-18 and ResNet-50 as backbone respectively. The number of feature channels and the scaling compared to the image are marked on the feature maps directly.}
  \label{fig5}
\end{figure*}

\subsection{Rethink of depth and width}
Following the backbone settings of the previous models, we adopt ResNet-50 as the encoder of  CTDNet, which is regarded as a standard version CTD-L. Then taking into account the three aspects of accuracy, parameters and speed, we design CTD-S and CTD-M with MobileNet-V2 and ResNet-18 from the perspective of width and depth respectively to explore the potential of efficient networks, which can respond to resource-constrained environments and the demand for fast processing of massive data.

The features extracted by different stages of backbone models are a mixture of low-level, mid-level and high-level features. Larger resolution features contribute less to performance but cost more computations, so we only use features of the last four stages $\{E^2,E^3,E^4,E^5\}$that have strides of $\{4, 8, 16, 32\}$ with respect to the input image.  The comparison of framework details on CTD-S, CTD-M and CTD-L is shown in \figref{fig5}. Specifically, compared with the conference version, our new proposed CTD-M based on ResNet-18 introduces the SAP structure in the decoder. which replace the original conv-1 module in the semantic path and improve the accuracy with little impact on the model parameters and speed. As to CTD-S which adopts MobileNet-V2 as backbone, the penultimate stage in the encoder contains two different blocks with the same spatial resolution. Since a narrow network cannot extract complex semantic features like deep networks, we remove the SAP used to extract semantic information in the decoder of CTD-S, and replace the downstream CAM with a common conv-3 module to further reduce the amount of model parameters. In contrast to CTD-S, CTD-L is able to obtain more complex features during encoding with its ResNet-50 backbone, so we can further mine $E^3$ and $E^2$ with SAP and SAM for extra semantic and spatial information respectively. In the following, we introduce each path and module by taking the framework of CTD-M as an example.

\subsection{Complementary Trilateral Decoder}
The proposed Complementary Trilateral Decoder (CTD) framework includes three branches: Semantic Path, Spatial Path and Boundary Path, which are derived from different levels of the shared encoder and are devised to confront the dilution of semantic context, loss of spatial structure and absence of boundary detail, respectively.

\textbf{Semantic Path:} Both semantic context and global context are helpful to locate salient objects accurately. However, one of the problems of the U-shape structure is that semantic information of high-level features will be gradually diluted when they are transmitted to lower layers in the top-down path. In addition, the receptive field of the lightweight networks is not large enough to capture global context information.

To solve these issues, we propose the Semantic Path to capture rich semantic context and global context, which can produce initial coarse saliency maps with accurate locations of salient objects. First, we embed a Global Average Pooling (GAP) layer on the tail of the backbone network, which can provide the maximum receptive field with the strongest global context. The output of global pooling is up-sampled and represented as $E^6$. Then we apply the proposed SAP structure to the latter two stages $E^4$ and $E^5$, which can enlarge the receptive field and capture multi-scale information efficiently. Finally, we design a simple Feature Fusion Module (FFM) to effectively fuse $E^4$, $E^5$ and$E^6$, which forms a partial U-shape structure (see \figref{fig3}). The Semantic Path can be formulized as:
\begin{equation}
  E_g^5 = {FFM}_1(SAP(E^5),Up(GAP(E^5))),
\end{equation}
\begin{equation}
  \mathcal{D}_{p1} = {FFM}_2(SAP(E^4),Up(E_g^5)),
\end{equation}
where $SAP$ and $Up$ represent the proposed SAP structure and up-sampling operations, respectively. $\mathcal{D}_{p1}$ represents the output of the Semantic Path.

After that, we introduce the detailed structure of FFM. To be specific, FFM receives two inputs $f_1$,$f_2$ and we adopt the multiplication operation to fuse these two features. Compared with addition and concatenation, the multiplication operation can avoid redundant information and suppress background noise. The fused features pass through two $3\times 3$ convolution layers to obtain more robust feature representation. The above process can be described as:
\begin{equation}
  FFM(f_1,f_2)=\mathcal{F}_{3\times 3}(\mathcal{F}_{3\times 3}(f_1\otimes f_2)),
\end{equation}
where $\mathcal{F}_{3\times3}$ represents $3\times 3$ convolution. Note that each convolution is followed by a batch normalization and a ReLU activation function.

\textbf{Spatial Path:} The Semantic Path captures rich semantic context and global context, while the Spatial Path is designed to preserve more spatial details. Spatial information is helpful to supplement structural details and generate complete segmentation results. However, spatial information is seriously lost after multiple down-samplings and cannot be recovered perfectly by integrating the hierarchical features from the encoder. Therefore, we propose the Spatial Path to learn more discriminative feature representation from spatial dimension.

The Spatial Path is drawn from mid-level features $E^3$ with large resolution (1/8 of the input size), which is beneficial to encode affluent spatial details. Specifically, we design a Spatial Attention Module (SAM) to refine features effectively (see \figref{fig4}). We first apply average channel pooling and maximum channel pooling that are performed pooling operations along the channel axis. These two generated single-channel spatial maps $S_{avg}$ and $S_{max}$ are concatenated. Then we compute a spatial attention map $M_{sa}$ by a $5\times 5$ convolution and sigmoid function. The spatial attention map $M_{sa}$ can re-weight the features $E^3$ from spatial dimension by element-wise multiplication. Finally, the weighted features $E_{sa}^3$ are fed into a $3\times 3$ convolution layer to squeeze the number of channels to 64. The Spatial Path can be formulized as:
\begin{equation}
  S_{avg} = {CP}_{avg}(E^3)=avg_{i\in [0,n-1]}(E_i^3),
\end{equation}
\begin{equation}
  S_{max} = {CP}_{max}(E^3)=max_{i\in [0,n-1]}(E_i^3),
\end{equation}
\begin{equation}
  M_{sa} = \sigma(\mathcal{F}_{5\times 5}(Concat(S_{avg}, S_{max}))),
\end{equation}
\begin{equation}
  \mathcal{D}_{p2} = \mathcal{F}_{3\times 3}(M_{sa}\otimes E^3) = \mathcal{F}_{3\times 3}(E_{sa}^3),
\end{equation}
where ${CP}_{avg}$ and ${CP}_{max}$ represent average channel pooling and maximum channel pooling operations, respectively. $E_i^3$ and $n$ denote the i-th channel of the feature map $E^3$ and the number of channels. $\mathcal{F}_{5\times 5}$ and Concat represent $5\times 5$ convolution and concatenation. $\sigma$ and $\otimes$ denote sigmoid function and element-wise multiplication. $\mathcal{D}_{p2}$ represents the output of the Spatial Path.

\textbf{Boundary Path:} The boundary contour can be regarded as the demarcation between the salient regions and the background. Boundary information is also helpful to segment salient objects accurately. However, we observe that saliency maps produced by many existing SOD methods based on the U-shape structure suffer from coarse boundaries. Therefore, we propose the Boundary Path to improve boundary quality by utilizing boundary contour explicitly. 
The Boundary Path is drawn from low-level features $E^2$, which preserves more boundary information due to larger resolution (1/4 of the input size). However, it is likely to bring noise and interference, such as the boundaries of non-salient regions. Therefore, we exploit high-level location information as guidance to help enhance salient boundary features and suppress non-salient boundary features with an extra edge supervision (see \figref{fig4}).
Specifically, we first apply a $1\times 1$ convolution to low-level features $E^2$ for channel compression. Then we up-sample the high-level features $E_g^5$ (see Eq. (1)) to the same size as $E^2$ by bilinear interpolation. Finally, we use the proposed FFM to fuse the local information and location information efficiently. In addition, we add an extra salient edge supervision to supervise the Boundary Path explicitly. The Boundary Path can be formulized as:
\begin{equation}
  \mathcal{D}_{p3} = FFM_{3}(\mathcal{F}_{1\times 1}(E^2),Up(E_g^5)),
\end{equation}
where $\mathcal{F}_{1\times 1}$ represents $1\times1$ convolution. Note that each convolution is followed by a batch normalization and a ReLU activation function. $\mathcal{D}_{p3}$ represents the output of the Boundary Path. From that we can know that Boundary Path can generate the boundary contours of salient objects, so our method can also be used for edge detection, which is incidental to our main task. Some visual examples can be found in \figref{fig4}.

\begin{figure}[t]
  \centering
  \includegraphics[width=1.0\columnwidth]{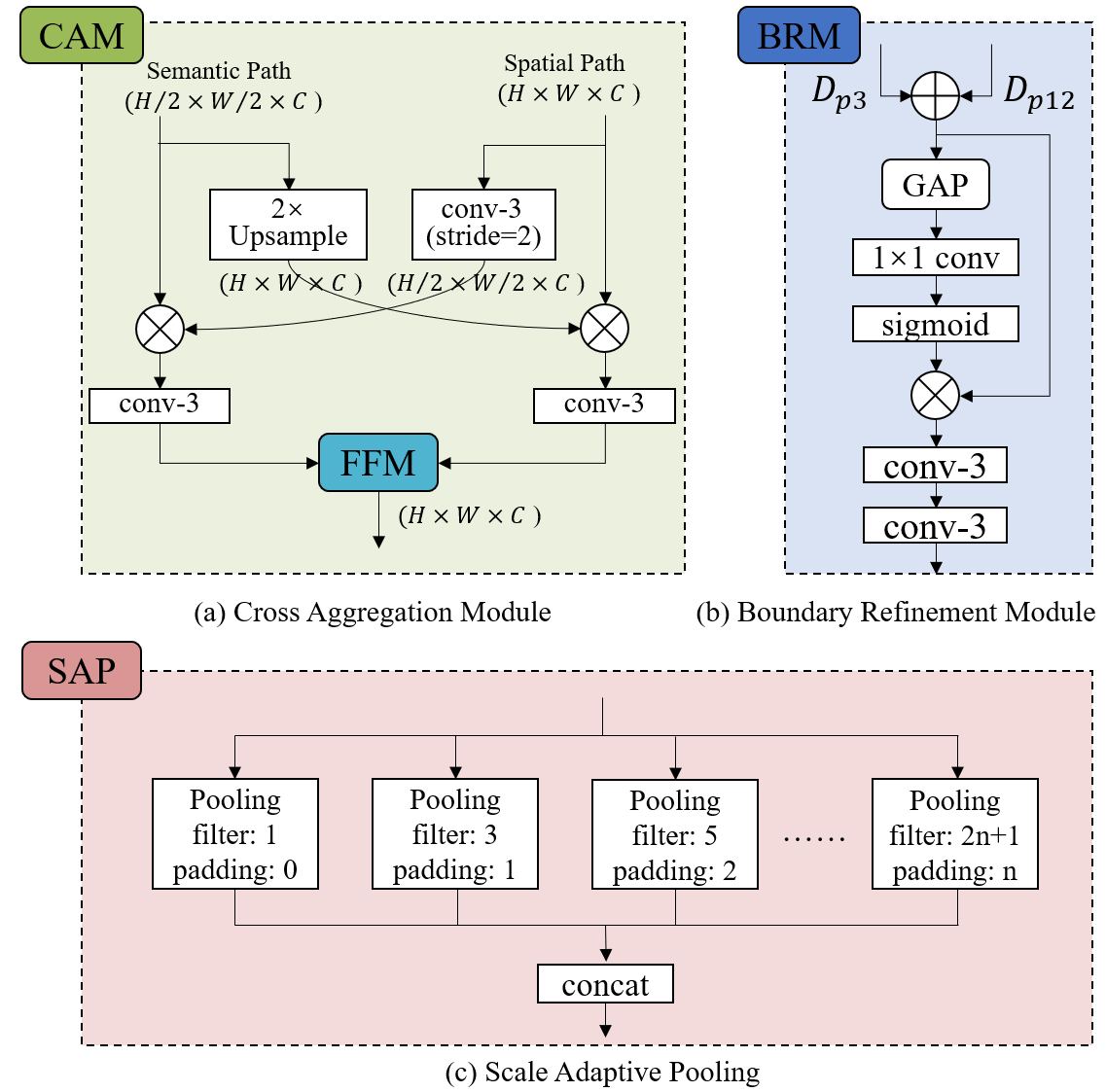} 
  \caption{The detailed structure of CAM, BRM and SAP}
  \label{fig6}
\end{figure}

\subsection{Interaction between Three Branches}
In order to make full use of the characteristics and complementarity of these three branches, we propose the Cross Aggregation Module (CAM) and Boundary Refinement Module (BRM) to gradually merge them according to “coarse-fine-finer” strategy, which significantly improves the region accuracy and boundary quality.

\textbf{Cross Aggregation Module:} The output of the Semantic Path $\mathcal{D}_{p1}$ contains rich semantic information with global context, which can produce initial coarse saliency maps with accurate locations of salient objects (see \figref{fig6}(a)). In contrast, the output of the Spatial Path $\mathcal{D}_{p2}$ preserves more spatial details. Both paths are complementary to each other, so we design a novel fusion module CAM to merge these two branches effectively, which can produce relative fine saliency maps with precise structures of salient objects (see \figref{fig8}(d)).
As \figref{fig6}(a) shows, the two inputs of CAM have different resolutions: $\mathcal{D}_{p1}\in\mathbb{R}^{\frac{H}{2}\times\frac{W}{2}\times C}$ and $\mathcal{D}_{p2}\in\mathbb{R}^{H\times W\times C}$. First, we perform the multi-scale transformation on each input. Specifically, we up-sample $\mathcal{D}_{p1}$to the same size as $\mathcal{D}_{p2}$ by bilinear interpolation and down-sample $\mathcal{D}_{p2}$ to the same size as $\mathcal{D}_{p1}$ by a $3\times 3$ convolution with stride 2, obtaining the corresponding features $\mathcal{D}_{p1}^\prime\in\mathbb{R}^{H\times W\times C}$ and $\mathcal{D}_{p2}^\prime\in\mathbb{R}^{\frac{H}{2}\times\frac{W}{2}\times C}$. Second, we perform cross aggregation on each scale by the multiplication operation and then apply a $3\times 3$ convolution respectively to adapt them, which can capture multi-scale information and promote interaction between two branches. Note that each convolution is followed by a batch normalization and a ReLU activation function. Finally, these two features $C_1\in\mathbb{R}^{\frac{H}{2}\times\frac{W}{2}\times C}$and $ C_2\in\mathbb{R}^{H\times W\times C}$ are fed into the proposed FFM to construct a comprehensive and powerful feature representation. The whole process can be described as:
\begin{equation}
  \mathcal{D}_{p1}^\prime =Up(\mathcal{D}_{p1}),\mathcal{D}_{p2}^\prime = {Down}_{3\times 3}(\mathcal{D}_{p2}),
\end{equation}
\begin{equation}
  C_1 = \mathcal{F}_{3\times 3}(\mathcal{D}_{p1}\otimes \mathcal{D}_{p2}^\prime), C_2 = \mathcal{F}_{3\times 3}(\mathcal{D}_{p2}\otimes \mathcal{D}_{p1}^\prime),
\end{equation}
\begin{equation}
  \mathcal{D}_{p12} = FFM_{4}(Up(C_1),C_2),
\end{equation}
where ${Down}_{3\times 3}$ denotes down-sampling operation using $3\times 3$ convolution with stride 2. $\mathcal{D}_{p12}$ represents the final output of the CAM.

\textbf{Boundary Refinement Module:} Although we obtain relatively fine saliency maps after merging the Semantic Path $\mathcal{D}_{p1}$ and the Spatial Path $\mathcal{D}_{p2}$, we can leverage the salient boundary information provided by the Boundary Path to further refine boundary. Therefore, we propose a fusion module BRM to merge $\mathcal{D}_{p12}$ and the Boundary Path $\mathcal{D}_{p3}$, which can generate final finer saliency maps with clear boundaries of salient objects (see \figref{fig8}(e)).

As \figref{fig6}(b) shows, we first concatenate the $\mathcal{D}_{p12}$ and $\mathcal{D}_{p3}$. Then we pool the fused features $B_f$ to generate a feature vector and compute an attention vector to guide the feature learning by a $1\times 1$ convolution and sigmoid function. This weight vector can re-weight the $B_f$ for feature selection and refinement by multiplication operation. Finally, the refined features $B_r$ are combined with $B_f$ and then pass through two $3\times 3$ convolution layers to further enhance feature representation. Note that each $3\times 3$ convolution is followed by a batch normalization and a ReLU activation function. The above process can be described as:
\begin{equation}
  B_f = Up(\mathcal{D}_{p12}) + \mathcal{D}_{p3},
\end{equation}
\begin{equation}
  B_r = B_f\otimes\sigma(\mathcal{F}_{1\times 1}(GAP(B_f))),
\end{equation}
\begin{equation}
  \mathcal{D}_{p123} = \mathcal{F}_{3\times 3}(\mathcal{F}_{3\times 3}(B_r+B_f)),
\end{equation}
$\mathcal{D}_{p123}$ represents the final output of the BRM.

\begin{figure*}[t]
  \centering
  \includegraphics[width=1.0\textwidth]{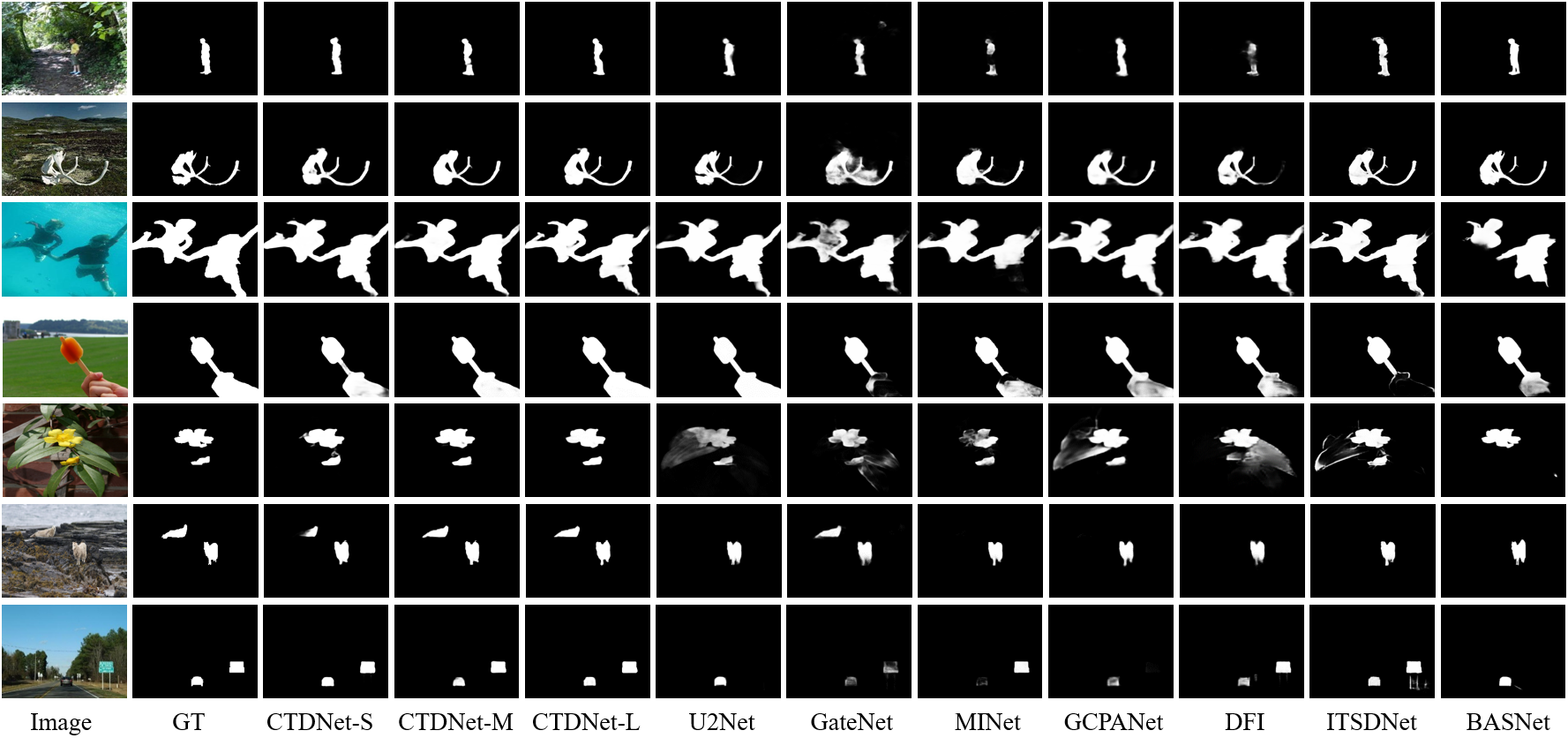} 
  \caption{Qualitative comparison of our model with existing state-of-the-art SOD models in some challenging scenarios}
  \label{fig7}
\end{figure*}

\subsection{Scale-Adaptive Pooling}
The receptive filed is of great significance for CNN, because the size of receptive filed can roughly indicate how much we can use context information. However, the empirical receptive fields of CNN are much smaller than the theoretical ones especially for high-level features. On the other hand, the multi-scale representation can help perceiving multi-scale objects, which is also indispensable for SOD. Some approaches have been proposed to enlarge the receptive field and capture multi-scale features, such as Pyramid Pooling Module (PPM)\cite{zhao2017pyramid} and Atrous Spatial Pyramid Pooling (ASPP)\cite{chen2017deeplab}, but these structures are computation demanding and memory consuming.

To this end, we propose a novel structure named Scale-Adaptive Pooling (SAP), as shown in \figref{fig6}(c). The scale-adaptive includes two aspects: receptive field scale-adaptive and spatial scale-adaptive. To be specific, we use $N$($N=4$) parallel pooling operations with stride $s = 1$, padding $p = n$ and filter $f = 2n + 1$, where $n = \{0, 1, · · · , N-1\}$. On the one hand, we use pooling of different kernel sizes to obtain multiple receptive fields, which can capture multi-scale context information and enlarge the receptive field to a certain extent. On the other hand, these pooling operations take a fixed stride of 1 and different paddings to keep the spatial size the same as input. Finally, the $N$ parallel branches are concatenated as the final output of SAP module.

  

Compared with PPM and ASPP, our proposed SAP has the following advantages: 1) Unlike ASPP that uses dilated convolutions with different dilation rates, SAP only uses pooling operations without increasing learnable parameters; 2) Different from PPM that requires frequent down-sampling and up-sampling, SAP does not change the spatial size of feature maps and reduces spatial information loss to a certain extent; 3) SAP can capture multi-scale context information from receptive fields of different sizes, which is more efficient and much faster.

\subsection{Loss Function}
In this paper, we adopt three loss functions: BCE loss \cite{de2005tutorial}, IoU loss \cite{mattyus2017deeproadmapper} and $L_1$ loss \cite{tibshirani1996regression}. BCE loss calculates the error for each pixel between the prediction mask and the ground truth, which is formulated as:
\begin{equation}
\begin{aligned}
    \ell_{bce}(P,G)= -\sum_{i=1}^{H}\sum_{j=1}^{W}[G(i,j)log(P(i,j)) \\
    +(1-G(i,j))log(1-P(i,j))],
\end{aligned}
\end{equation}
where $P(i,j)$ and $G(i,j)$ represent the pixel of prediction mask (P) and the ground truth (G) at location $(i,j)$ in an image. H and W are the height and width of the image, respectively. IoU loss is used to measure the similarity of structure instead of focusing on single pixel. We adopt the following form:
\begin{small}
  \begin{equation}
    \begin{aligned}
      \ell_{iou}\left(P,G\right)=1- 
      \frac{\sum_{i=1}^{H}\sum_{j=1}^{W}G\left(i,j\right)P\left(i,j\right)}{\sum_{i=1}^{H}\sum_{j=1}^{W}\left[G\left(i,j\right)+P\left(i,j\right)-G\left(i,j\right)p\left(i,j\right)\right]},
    \end{aligned}
  \end{equation}
\end{small}
$L_1$ loss is used to measure the minimum absolute error for each pixel between the prediction mask and the ground truth, which is formulated as:
\begin{equation}
  \begin{aligned}
    \ell_1\left(P,G\right)=\sum_{i=1}^{H}\sum_{j=1}^{W}\left|G\left(i,j\right)-P\left(i,j\right)\right|,
  \end{aligned}
\end{equation}

As described above, our model is deeply supervised with six outputs. All outputs pass through a $3\times 3$ convolution and sigmoid function to convert the feature maps to the corresponding single-channel prediction masks. For $\mathcal{D}_{p123}$, $\mathcal{D}_{p12}$, $\mathcal{D}_{p1}$, $E_g^5$ and $E^6$, we use three loss functions together to supervise these five saliency maps (see Eq. (17)), while for $\mathcal{D}_{p3}$, we use BCE loss and $L_1$ loss to supervise the boundary prediction mask ($P_b$). Note that the ground truth of salient boundary ($G_b$) can be easily obtained from the ground truth of salient objects.
\begin{equation}
  \begin{aligned}
    \ell_s\left(P,G\right)\ =\ell_{iou}\left(P,G\right)\ +\beta\ell_{bce}\left(P,G\right)+\gamma \ell_1\left(P,G\right),       
  \end{aligned}
\end{equation}
\begin{equation}
  \begin{aligned}
    \ell_b\left(P_b,G_b\right)=\frac{1}{2}\left(\ell_{bce}\left(P_b,G_b\right)+l_1\left(P_b,G_b\right)\right),
  \end{aligned}
\end{equation}
where $\beta$ and $\gamma$ are hyperparameters to balance the weight between the three loss functions, so that the network can achieve better performance. In our paper, the parameter $\beta$ is set to 0.6. The total loss function is denoted as follows: 
\begin{equation}
  \begin{aligned}
    \mathcal{L}\left(P,P_b,G,G_b\right)\ =l_b(P_b,G_b)\ +\sum_{k=1}^{5}{\alpha_kl_s\left(P^k,G\right),}                
  \end{aligned}
\end{equation}
where $\alpha_k$ denotes the weight of the $k-th$ loss term.

\begin{table*}[t]
\caption{ Quantitative comparisons with state-of-the-art SOD models on five benchmarks in terms of parameters, speed,$mF_{\beta }$,$\mathrm{MAE}$,$E_{m}$. The speed data marked with $\dag$ are uniformly re-measured on Titan Xp GPU in the same environment. The best two results are shown in red and green, respectively.}
\renewcommand{\arraystretch}{1}
\setlength\tabcolsep{1.5pt}
\resizebox{\textwidth}{!}{
\begin{tabular}{l|c|c|ccc|ccc|ccc|ccc|ccc}
\toprule
\multirow{2}{*}{\textbf{Method}} & \textbf{Params} &\textbf{Speed} &\multicolumn{3}{c|}{\textbf{ECSSD}} &\multicolumn{3}{c|}{\textbf{PASCAL-S}} &\multicolumn{3}{c|}{\textbf{DUTS-TE}} &\multicolumn{3}{c|}{\textbf{HKU-IS}} &\multicolumn{3}{c}{\textbf{DUT-OMRON}}\\
\cline{4-18}

~ &\textbf{(M)} & \textbf{(FPS)} 
& $mF_{\beta }$ &$\mathrm{MAE}$ & $E_{m}$
& $mF_{\beta }$ &$\mathrm{MAE}$ & $E_{m}$
& $mF_{\beta }$ &$\mathrm{MAE}$ & $E_{m}$
& $mF_{\beta }$ &$\mathrm{MAE}$ & $E_{m}$
& $mF_{\beta }$ &$\mathrm{MAE}$ & $E_{m}$ \\
\toprule
\multicolumn{18}{c}{\textbf{VGG/ResNet-based Models}} \\
\midrule
\textbf{C2SNet$_{18}$}                 & 158.86                                                                          & 30                                                                               & .864                        & .055                        & .914                        & .758                        & .080                        & .839                        & .716                        & .063                        & .846                        & .851                        & .048                        & .927                        & .683                        & .072                        & .829                        \\
\textbf{RAS$_{18}$}                    & 21.23                                                                           & 35                                                                               & .889                        & .056                        & .914                        & .777                        & .101                        & .829                        & .751                        & .059                        & .861                        & .871                        & .045                        & .929                        & .713                        & .062                        & .846                        \\
\textbf{PiCANet$_{18}$}                & 38.32                                                                           & 7                                                                                & .885                        & .046                        & .910                        & .789                        & .077                        & .828                        & .749                        & .054                        & .852                        & .870                        & .042                        & .934                        & .710                        & .068                        & .834                        \\
\textbf{BMPM$_{18}$}                   & 75.07                                                                           & 22                                                                               & .868                        & .045                        & .914                        & .758                        & .073                        & .836                        & .745                        & .049                        & .860                        & .871                        & .039                        & .937                        & .692                        & .064                        & .837                        \\
\textbf{PAGE$_{19}$}                   & 47.40                                                                           & 25                                                                               & .906                        & .042                        & .920                        & .806                        & .075                        & .841                        & .777                        & .052                        & .869                        & .882                        & .037                        & .940                        & .736                        & .062                        & .853                        \\
\textbf{AFNet$_{19}$}                  & 35.99                                                                           & 26                                                                               & .908                        & .042                        & .918                        & .820                        & .070                        & .850                        & .793                        & .046                        & .879                        & .888                        & .036                        & .942                        & .738                        & .057                        & .853                        \\
\textbf{BANet$_{19}$}                & 56.02                                                                           & 13                                                                                & .923                        & .035                        & \color{green}{.928}                       & .823                        & .069                        & .852                        & .815                        & .040                        & .892                        & .900                        & .032                        & .950                        & .746                        & .059                        & .860                        \\
\textbf{EGNet$_{19}$}                  & 111.78                                                                          & 8                                                                                & .920                        & .037                        & .927                        & .817                        & .073                        & .848                        & .815                        & .039                        & .891                        & .901                        & .031                        & .950                        & .755                        & .053                        & .867
\\
\textbf{SCRN$_{19}$}                   & 25.32                                                                           & 32                                                                               & .918                        & .038                        & .926                        & .826                        & .064                        & .857                        & .809                        & .040                        & .888                        & .896                        & .034                        & .949                        & .746                        & .056                        & .863                        \\
\textbf{PoolNet$_{19}$}                & 68.16                                                                           & 18                                                                               & .915                        & .039                        & .924                        & .815                        & .074                        & .848                        & .809                        & .040                        & .889                        & .899                        & .032                        & .949                        & .747                        & .056                        & .863                        \\
\textbf{CPD$_{19}$}                    & 47.97                                                                           & \color{green}{62}                                                                               & .917                        & .037                        & .925                        & .820                        & .070                        & .849                        & .805                        & .043                        & .887                        & .891                        & .034                        & .944                        & .747                        & .056                        & .866                        \\
\textbf{BASNet$_{19}$}                 & 87.03                                                                           & 25                                                                               & .880                        & .037                        & .921                        & .771                        & .075                        & .846                        & .791                        & .048                        & .884                        & .895                        & .032                        & .946                        & .756                        & .056                        & .869
\\
\textbf{GateNet$_{20}$}                & -                                                                               & -                                                                                & .916                        & .040                        & .924                        & .819                        & .067                        & .851                        & .807                        & .040                        & .889                        & .899                        & .033                        & .949                        & .746                        & .055                        & .862                        \\
\textbf{U2Net$_{20}$}                  &  46.21                                                     & 30                                                                               & .892                        &\color{green}{ .033}                       & .924                        & .770                        & .073                        & .842                        & .792                        & .045                        & .886                        & .896                        & .031                        & .948                        & .761                        & .054                        & .871                        \\
\textbf{DFI$_{20}$}                    & 29.57                                                                           & 57                                                                               & .920                        & .038                        & .924                        & .830                        & .064                        & .855                        & .814                        & .039                        & .892                        & .901                        & .031                        & .951                        & .752                        & .055                        & .865                        \\
\textbf{GCPANet$_{20}$}                & 67.05                                                                           & 50                                                                               & .919                        & .035                        & .920                        & .827                        & \color{green}{ .061}
& .847                        & .817                        & .038                        & .891                        & .898                        & .031                        & .949                        & .748                        & .056                        & .860                        \\
\textbf{ITSDNet$_{20}$}                & \color{green}{ 26.55}                                                                           & 43                                                                               & .895                        & .035                        & .927                        & .785                        & .071                        & .850                        & .804                        & .041                        & .895                        & .899                        & .031                        & .952                        & .756                        & .061                        & .863                        \\ 
\textbf{MINet$_{20}$}                  & 162.38                                                                          & 31                                                                               & .924                        & \color{green}{.033}            & .927                        & .829                        &  .063                        & .851                        & .828                        &  .037                        & .898                        & .909                        & .029                       & .953                        & .755                        & .055                        & .865
\\
\textbf{CII$_{21}$}
&24.48
&50
&.929
&\color{green}{.033}
&.926
&.841
&\color{green}{.061}
&.857
&.834
&.037
&.900
&.915
&.029
&.953
&.768
&.054
&.872
\\
\textbf{EDN$_{22}$}
&42.85
&51.7
&\color{red}{.932}
&\color{red}{.032}
&\color{red}{.929}
&\color{red}{.847}
&\color{green}{.061}
&\color{red}{.864}
&\color{green}{.851}
&\color{green}{.035}
&\color{green}{.908}
&\color{green}{.919}
&\color{green}{.026}
&\color{green}{.956}
&\color{red}{.783}
&\color{red}{.048}
& \color{green}{.876}
\\

\textbf{CTD-L (Ours)} &
\color{red}{26.48} &
\color{red}{84} &

\color{green}{.931} &
\color{red}{.032} &
.925 &

\color{green}{.845} &
\color{red}{.059} &
\color{green}{.860} &

\color{red}{.863} &
\color{red}{.032} &
\color{red}{.914} &

\color{red}{.922} &
\color{red}{.025} &
\color{red}{.957} &

\color{red}{.783} &
\color{green}{.049} &
\color{red}{.878} \\

\toprule
\multicolumn{18}{c}{\textbf{Lightweight Models}}                                                                            \\ 
\midrule
\textbf{SAMNet$_{21}$}
&\color{red}{1.33}
& $28^{\dag}$
&.891
&.050
&.911
&.778
&.090
&.823
&.745
&.058
&.849
&.871
&.045
&.934
&.717
&.065
&.840
\\
\textbf{CII (ResNet-18)$_{21}$}
&11.89
&$100^{\dag}$
&.915
&.039
&.921
&.820
&.067
&.848
&.814
&.043
&\color{green}{.890}
&.906
&.032
&.949
&.745
&.058
&.860
\\
\textbf{EDN-Lite$_{22}$}
&1.8
&$44^{\dag}$
&\color{green}{.916}
&.042
&.919
&.820
&.072
&\color{green}{.853}
&.809
&.045
&.888
&.901
&.034
&.945
&.748
&.057
&.857
\\
  \textbf{CTD-S (Ours)} &
  \color{green}{1.7} &
  \color{green}{125} &

  .915 &
  \color{green}{.038} &
  \color{red}{.924} &

  \color{green}{.826} &
  \color{green}{.066} &
  .851 &
  
  \color{green}{.816} &
  \color{green}{.041} &
  \color{green}{.890} &
  
  \color{green}{.907} &
  \color{green}{.029} &
  \color{green}{.951} &

  \color{green}{.758} &
  \color{green}{.056} &
  \color{green}{.863} \\

  \textbf{CTD-M (Ours)} &
  12.6 &
  \color{red}{158} &

  \color{red}{.923} &
  \color{red}{.035} &
  \color{green}{.922} &

  \color{red}{.831} &
  \color{red}{.065} &
  \color{red}{.858} &

  \color{red}{.835} &
  \color{red}{.038} &
  \color{red}{.900} &

  \color{red}{.915} &
  \color{red}{.028} &
  \color{red}{.954} &

  \color{red}{.767} &
  \color{red}{.054} &
  \color{red}{.869} \\

\bottomrule

\end{tabular}
}
\label{tab1}
\end{table*}

\section{Experiments}

\subsection{Datasets} 
We conduct experiments on five standard benchmark datasets: ECSSD \cite{yan2013hierarchical} , PASCAL-S \cite{li2014secrets}, DUTS \cite{wang2017learning}, HKU-IS \cite{li2015visual} and DUT-OMRON \cite{yang2013saliency}, and the detailed introduction is provided as follows: ECSSD contains 1,000 natural images with many semantically meaningful and complex structures in their ground-truth segmentation. PASCAL-S consists of 850 natural images that are carefully selected from the PASCAL VOC dataset. DUTS is currently the largest SOD dataset, which contains 10,553 images for training (DUTS-TR) and 5,019 images for testing (DUTS-TE). HKU-IS includes 4,447 challenging images and most of them have multiple disconnected salient objects, overlapping image boundaries or low color contrast. DUT-OMRON has 5,168 high quality images, which have one or more salient objects and relatively cluttered background. 

\subsection{Evaluation Metrics}
To quantitatively evaluate the performance, we adopt three evaluation metrics: Mean Absolute Error (MAE), F-measure and E-measure \cite{fan2018enhanced}. MAE measures the pixel-wise average absolute difference between the prediction mask and ground truth, which is formulated as:
\begin{equation}
  \begin{aligned}
    MAE=\frac{1}{W\times H}\sum_{i=1}^{H}\sum_{j=1}^{W}\left|P\left(i,j\right)-G\left(i,j\right)\right|,           
  \end{aligned}
\end{equation}
where P and G represent the prediction mask and the corresponding ground truth, respectively. H, W are the height and width of the image. The smaller MAE indicates better performance. Another metric F-measure ($F_\beta$) takes both precision and recall into account, which is defined as:
\begin{equation}
  \begin{aligned}
    F_\beta=\frac{\left(1+\beta^2\right)\cdot Precision\cdot Recall}{\beta^2\cdot Precision+Recall}          
  \end{aligned}
\end{equation}
where $\beta^2$ is set to 0.3 to emphasize the precision over recall. Larger F-measure indicates better performance. We choose the max F-measure ($mF_\beta$) in our paper. In addition, E-measure ($E_m$) combines local pixel values with the image-level mean value to jointly evaluate the similarity between the prediction mask and the ground truth.

\begin{figure}[t!]
  \centering
  \includegraphics[width=0.97\columnwidth]{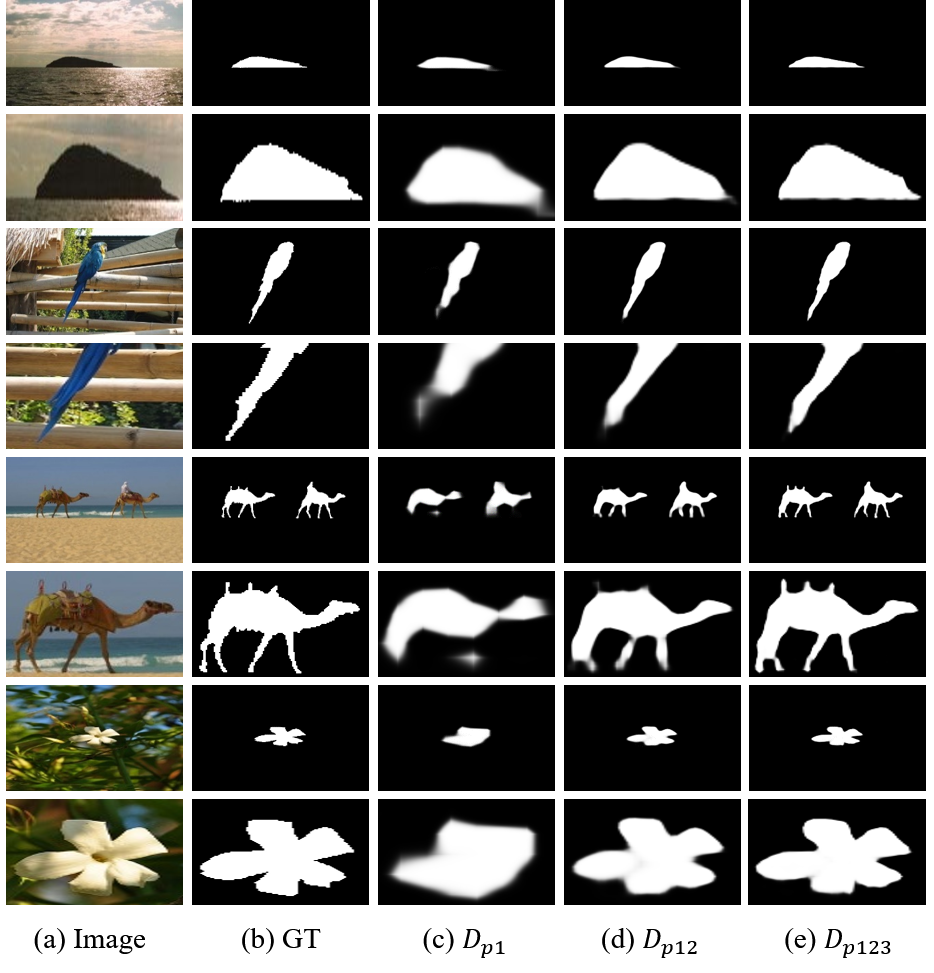} 
  \caption{The saliency maps from different locations of our network. Each example contains two rows and the second row of each example denotes the zoom-in view of salient object. The results conform to “coarse-fine-finer” predictions along with the gradual combination of these three branches, which demonstrates the complementarity of these three branches.}  
  \label{fig8}
\end{figure}

\subsection{Implementation Details}
We implement our proposed method by PyTorch and conduct experiments on one NVIDIA 1080Ti GPU. We use MobileNet-V2, ResNet-18 and ResNet-50 pre-trained on ImageNet as backbone networks, respectively. We choose DUTS-TR as training dataset as used in \cite{liu2018picanet},\cite{wang2018detect} and evaluate our model on other datasets. In the training period, all training images are resized to 352×352 with random cropping and random horizontal flipping to feed into the proposed model. We use stochastic gradient descent (SGD) optimizer with momentum of 0.9 and weight decay of 5e-4 to train our model. The batch size is set to 32 and maximum epoch is set to 48. We adopt the warm-up and linear decay learning rate strategy with the maximum learning rate 5e-3 for pre-trained backbone and 5e-2 for the rest of network. During the inference period, each image is simply resized to 352×352 and then fed into our model to predict saliency map without any post-processing (e.g., CRF \cite{krahenbuhl2011efficient}).

\begin{table*}[t]
  \scriptsize
  \caption{The ablation study of our proposed components. The backbone network takes ResNet-18 as an example. By adding each module gradually, our model achieves the best performance.}
  \renewcommand{\arraystretch}{1.2}
  \setlength\tabcolsep{1.5pt}
  \centering
  \resizebox{\textwidth}{!}{
  \begin{tabular}{c|ccc|cc|cc|c|c|ccc|ccc}
  \toprule
  \multirow{2}{*}{\textbf{Base}} 
  & \multicolumn{3}{c|}{\textbf{Semantic Path}}
  & \multicolumn{2}{c|}{\textbf{Spatial Path}}
  & \multicolumn{2}{c|}{\textbf{Boundary Path}}
  & \textbf{Params} 
  & \textbf{Speed} 
  &\multicolumn{3}{c|}{\textbf{DUTS-TE}} 
  &\multicolumn{3}{c}{\textbf{DUT-OMRON}}\\
  
  \cline{11-16}
  \cline{2-8}
  ~ &FFM &SAP &Global &SAM &CAM &FFM &BRM &\textbf{(M)} & \textbf{(FPS)} 
  & $mF_{\beta }$ &$\mathrm{MAE}$ & $E_{m}$
  & $mF_{\beta }$ &$\mathrm{MAE}$ & $E_{m}$ \\
  \midrule
  \checkmark &~ &~ &~ &~ &~ &~ &~ &11.352 &270 &.797 &.046 &.880 &.724 &.064 &.843 \\
  \checkmark &\checkmark &~ &~ &~ &~ &~ &~ &11.389 &260 &.803 &.046 &.882 &.733 &.064 &.847 \\
  \checkmark &\checkmark &\checkmark &~ &~ &~ &~ &~ &12.127 &240 &.806 &.044 &.882 &.742 &.06 &.855 \\
  \checkmark &\checkmark &\checkmark &\checkmark &~ &~ &~ &~ &12.234 &220 &.811 &.043 &.886 &.744 &.06 &.855 \\
  \checkmark &\checkmark &\checkmark &\checkmark &\checkmark &~ &~ &~ &12.300 &210 &.817 &.042 &.89 &.751 &.059 &.857 \\
  \checkmark &\checkmark &\checkmark &\checkmark &\checkmark &\checkmark &~ &~ &12.448 &190 &.821 &.041 &.891 &.753 &.057 &.863 \\
  \checkmark &\checkmark &\checkmark &\checkmark &\checkmark &\checkmark &\checkmark &~ &12.522 &180 &.826 &.040 &.895 &.754 &.056 &.863 \\
  \checkmark &\checkmark &\checkmark &\checkmark &\checkmark &\checkmark &\checkmark &\checkmark  &12.612 &160 &.835 &.038 &.900 &.767 &.054 &.869 \\
  
  \bottomrule
  \end{tabular}
  }
  \label{tab2}
\end{table*}

\subsection{Comparison results}
To prove the effectiveness of our method, we compare with 18 state-of-the-art SOD models, including C2SNet \cite{li2018contour}, RAS \cite{chen2018reverse}, PiCANet \cite{liu2018picanet}, BMPM \cite{zhang2018bi}, BANet \cite{su2019banet}, EGNet \cite{zhao2019egnet}, SCRN \cite{wu2019stacked}, PoolNet \cite{liu2019simple}, PAGE \cite{wang2019salient}, AFNet \cite{feng2019attentive}, CPD \cite{wu2019cascaded}, BASNet \cite{qin2019basnet}, GateNet \cite{zhao2020suppress}, DFI \cite{liu2020dynamic}, ITSDNet \cite{zhou2020interactive}, GCPANet \cite{chen2020global}, MINet \cite{pang2020multi}, U2Net \cite{qin2020u2}, CII \cite{liu2021rethinking}, EDN \cite{wu2022edn} and SAMNet \cite{liu2021samnet}. For a fair comparison, we use saliency maps released by the authors and evaluate them with the same Matlab code.

\textbf{Qualitative Comparison.} To intuitively show the advantages of our three models, we provide some visual examples of various SOD models, as shown in \figref{fig7}. We can observe that our models CTD-S, CTD-M and CTD-L can generate more complete and more accurate segmentation results than other counterparts. It can handle various challenging scenarios, such as multiple salient objects (row 3, 6 and 7), fine structure (row 2, 4 and 5), cluttered backgrounds (row 1 and 5) and small objects (row 6 and 7). In addition, we do not use any post-processing to obtain these results. Therefore, our model shows its effectiveness and robustness in processing complicated images.

\textbf{Quantitative Comparison.} \tabref{tab1} shows the quantitative results on five popular datasets in terms of $mF_\beta$, MAE and $E_m$. To facilitate the practical application in different environments, we adopt MobileNet-V2, ResNet-18 and ResNet-50 as backbones respectively and name our model CTD-S, CTD-M and CTD-L accordingly. Compared with VGG/ResNet-based models, CTD-L obtains the best performance under almost all evaluation metrics on five benchmarks. More importantly, our models CTD-S and CTD-M outperform other lightweight models and achieve comparable or even better performance than VGG/ResNet-based models. In addition, we also list the parameters and speed of each method to measure efficiency. The MobileNet-based CTD-S that can run at a 125 FPS speed only has 1.7M parameters, which is more than five times compressed compared to the most existing model parameters; the ResNet-based CTD-M only has 12.6M parameters and runs at a speed of 158 FPS on one GTX 1080Ti GPU for 352×352 input images, which surpasses the existing approaches by a large margin. Moreover, CTD-L runs at a 84 FPS speed with 26.5M parameters, which is much smaller and faster than the existing VGG/ResNet-based SOD methods. It should be noted that although CTD-S has fewer parameters than CTD-M, it is not as fast as CTD-M due to the depthwise separable convolutions in its backbone\cite{orsic2019defense}. In conclusion, our model achieves a favorable trade-off between speed and accuracy, which clearly demonstrates its superiority and efficiency.

\begin{table}[t!]
  \caption{The complementarity of these three branches. $\mathcal{D}_{p1}$,$\mathcal{D}_{p12}$ and $\mathcal{D}_{p123}$ denote the Semantic Path, the combination
  of both Semantic Path and Spatial Path, the combination of these three branches, respectively}
  \renewcommand{\arraystretch}{1.2}
  \setlength\tabcolsep{4pt}
  \resizebox{1.0\columnwidth}{!}{
  \begin{tabular}{c|ccc|ccc}
  \toprule
  \multirow{2}{*}{\textbf{Merge}} 
  & \multicolumn{3}{c|}{\textbf{DUTS-TE}}
  & \multicolumn{3}{c}{\textbf{DUT-OMRON}}\\
  
  \cline{2-7}
  ~
  & $mF_{\beta }$ &$\mathrm{MAE}$ & $E_{m}$
  & $mF_{\beta }$ &$\mathrm{MAE}$ & $E_{m}$ \\
  \midrule
  $\mathcal{D}_{p1}$ & .766 & .049 & .872 & .712 & .063 & .848 \\
  $\mathcal{D}_{p12}$ & .819 & .040 & .894 & .754 & .056 & .866 \\
  $\mathcal{D}_{p123}$ &.835 &.038 &.900 &.767 &.054 &.869 \\
  \bottomrule
  \end{tabular}
  }
  \label{tab3}
\end{table}

\begin{table}[t!]
  \caption{The improvements of accuracy with SAP module compared to the conference version}
  \renewcommand{\arraystretch}{1.2}
  \setlength\tabcolsep{4pt}
  \resizebox{1.0\columnwidth}{!}{
  \begin{tabular}{c|ccc|ccc}
  \toprule
  \multirow{2}{*}{\textbf{Model}} 
  & \multicolumn{3}{c|}{\textbf{DUTS-TE}}
  & \multicolumn{3}{c}{\textbf{DUT-OMRON}}\\
  
  \cline{2-7}
  ~
  & $mF_{\beta }$ &$\mathrm{MAE}$ & $E_{m}$
  & $mF_{\beta }$ &$\mathrm{MAE}$ & $E_{m}$ \\
  \midrule
  CTDNet(conference) & .853 & .034 & .909 & .779 & .052 & .875 \\
  CTDNet with SAP & .856 & .033 & .912 & .781 & .050 & .876 \\
  \bottomrule
  \end{tabular}
  }
  \label{tab4}
\end{table}

\subsection{Ablation Study}
Firstly, we investigate the complementarity of these three branches. Secondly, we verify the effectiveness of our proposed components. Lastly, we validate the improvements of accuracy with SAP module compared to the conference version. All experiments are conducted on DUTS-TE and DUT-OMRON datasets.

\textbf{The complementarity of these three branches.} To demonstrate the complementarity and necessity of these three branches, we conduct experiments both qualitatively and quantitatively. As shown in \tabref{tab3}, when merging the Semantic Path $\mathcal{D}_{p1}$ and Spatial Path $\mathcal{D}_{p2}$, the performance can be greatly improved. Moreover, the performance can be further boosted by merging $\mathcal{D}_{p12}$ and $\mathcal{D}_{p3}$, which benefits from the salient boundary information provided by the Boundary Path $\mathcal{D}_{p3}$. In addition, we visualize some examples in \figref{fig8}. Each example contains two rows and the second row of each example denotes the zoom-in view of salient object. As we can see, the produced saliency maps conform to “coarse-fine-finer” predictions along with the gradual combination of these three branches. Column 3 represents initial coarse saliency maps produced by $\mathcal{D}_{p1}$ with accurate locations of salient objects. Column 4 represents relatively fine saliency maps produced by $\mathcal{D}_{p12}$ with precise structures of salient objects. Column 5 represents final finer saliency maps produced by $\mathcal{D}_{p123}$ with clear boundaries of salient objects. Obviously, experimental results verify the complementarity and necessity of these three branches.

\textbf{The effectiveness of our proposed components.} To demonstrate the effectiveness of our proposed components, we conduct ablation experiments by gradually adding them. First, we replace all the proposed fusion modules with simple addition operation followed by the $3\times 3$ convolution to construct a baseline network, which still maintains three branches in the decoder. Second, we gradually add the FFM, SAP and global context for the Semantic Path and Boundary Path. Then we add the SAM in the Spatial Path. Finally, we use the proposed fusion modules CAM and BRM to merge these three branches. As shown in \tabref{tab2}, our method achieves the best performance when all modules are contained, which demonstrates the effectiveness and necessity of each module.

\textbf{The improvements of accuracy with SAP module.} To explore the extent to which the newly proposed module SAP improves the accuracy compared to the conference version, we conduct ablation experiments by only adding SAP modules into the CTDNet of the conference version. As shown in \tabref{tab4}, the SAP module effectively improves the accuracy of the model and helps the model obtain the best performance against state-of-the-art methods. Note that although the newly added module SAP has slightly slowed down the speed compared with conference version, the overall efficiency of our new proposed model is still excellent compared with other counterparts and the accuracy has been effectively improved.

\section{Conclusion}
In this paper, we first reveal that existing salient object detection methods adopt deeper and wider networks to achieve better performance, resulting in imbalance in accuracy, parameters and speed. To this end, we design a lightweight framework while maintaining satisfying competitive accuracy. Specifically, after analyzing the drawbacks of the U-shape structure, we propose a novel trilateral decoder framework by decoupling the U-shape structure into three complementary branches, which are devised to confront the dilution of semantic context, loss of spatial structure and absence of boundary detail, respectively.  Along with the fusion of three branches, the coarse segmentation results are gradually refined in structure details and boundary quality. Taking the advantages of pooling operation, we propose the Scale-Adaptive Pooling to obtain multi-scale receptive filed without additional learnable parameters. On the premise of inheriting this framework, we explore the maximum potential of shallow and narrow network and provide three versions: CTD-S (1.7M, 125FPS), CTD-M(12.6M, 158FPS) and CTD-L (26.5M, 84FPS) to facilitate the practical application in different environments. Experiments demonstrate that our proposed method performs better than state-of-the-art methods on five benchmarks, which achieves a favorable trade-off between efficiency and performance.

\section{Acknowledgement}
This work is partially supported by the National Natural Science Foundation of China under the Grant 62132002 and Grant 62102206.

{\small
\bibliographystyle{IEEEtran}
\bibliography{ctd_ref}
}

\vfill
\end{document}